%% file: root.tex
\documentclass[letterpaper, 10 pt, conference]{ieeeconf}  

\IEEEoverridecommandlockouts                              

\usepackage{graphicx} 
\usepackage{amsmath} 
\usepackage{amssymb}  
\usepackage{cite}
\usepackage{amsmath}
\usepackage{color}
\usepackage{booktabs}
\usepackage{color}
\usepackage{xspace}

\newcommand\projectname{ReSPEC\xspace}

\title{\LARGE \bf
ReSPEC: A Framework for Online Multispectral Sensor Reconfiguration in Dynamic Environments
}

\author{Anonymous Authors}

\author{Yanchen Liu$^{1}$, Yuang Fan$^{1}$, Minghui Zhao$^{1}$ and Xiaofan Jiang$^{1}$
\thanks{$^{1}$Yanchen Liu, Minghui Zhao, and Xiaofan Jiang are with the Department of Electrical Engineering, Columbia University, USA
         {\tt\small \{yl4189, yf2676, mz2866\}@columbia.edu, jiang@ee.columbia.edu}}%
}

\begin{document}

\maketitle
\thispagestyle{empty}
\pagestyle{empty}

\begin{abstract}
Multi-sensor fusion is central to robust robotic perception, yet most existing systems operate under static sensor configurations, collecting all modalities at fixed rates and fidelity regardless of their situational utility. This rigidity wastes bandwidth, computation, and energy, and prevents systems from prioritizing sensors under challenging conditions such as poor lighting or occlusion. Recent advances in reinforcement learning (RL) and modality-aware fusion suggest the potential for adaptive perception, but prior efforts have largely focused on re-weighting features at inference time, ignoring the physical cost of sensor data collection. We introduce a framework that unifies sensing, learning, and actuation into a closed reconfiguration loop. A task-specific detection backbone extracts multispectral features (e.g. RGB, IR, mmWave, depth) and produces quantitative contribution scores for each modality. These scores are passed to an RL agent, which dynamically adjusts sensor configurations, including sampling frequency, resolution, sensing range, and etc., in real time. Less informative sensors are down-sampled or deactivated, while critical sensors are sampled at higher fidelity as environmental conditions evolve. We implement and evaluate this framework on a mobile rover, showing that adaptive control reduces GPU load by 29.3\% with only a 5.3\% accuracy drop compared to a heuristic baseline. These results highlight the potential of resource-aware adaptive sensing for embedded robotic platforms.
\end{abstract}
\input{sections/1-intro}

\input{sections/2-related}
\input{sections/3-method}

\input{sections/4-experiment}
\input{sections/5-conclusion}





\section*{Acknowledgment}
This research was partially supported by COGNISENSE, one of seven centers in JUMP 2.0, a Semiconductor Research Corporation (SRC) program sponsored by DARPA, as well as the National Science Foundation under Grant Number CNS-1943396. The views and conclusions contained here are those of the authors and should not be interpreted as necessarily representing the official policies or endorsements, either expressed or implied, of Columbia University, NSF, SRC, DARPA, or the U.S. Government or any of its agencies.

\bibliographystyle{IEEEtran}
\bibliography{references}

\end{document}

%% file: sections/1-intro.tex
\section{Introduction}

Multi-sensor fusion has become a cornerstone of robotic perception, enabling systems to leverage complementary modalities such as RGB cameras, infrared sensors, LiDAR, and mmWave radar to operate reliably across diverse environments~\cite{ruan_survey_2025, qin_vins-mono_2018, liu_bevfusion_2024, shi_radar_2024, lin_rcbevdet_2024, brodermann_muses_2025}. Despite impressive advances in fusion architectures, most existing systems rely on static sensor configurations, collecting and processing all sensor data at fixed rates and fidelity regardless of their situational utility~\cite{palladin_samfusion_2025, zhang_tfdet_2025, park_resilient_2025, cho_cocoon_2024, qu_active_2017}. This approach results in inefficiencies: redundant data from less informative sensors waste bandwidth, computation, and energy, while important sensors cannot be dynamically prioritized under challenging conditions such as poor lighting or occlusion.

Recent progress in reinforcement learning (RL) and modality-aware fusion has opened opportunities for \textit{adaptive perception}, where sensor usage is selectively modulated to optimize both performance and resource efficiency. However, most prior works implement adaptivity only in the feature space, for example by re-weighting or gating modality embeddings at inference time~\cite{cao_multi-modal_2023, he_ecient_2024}. Such approaches overlook an important dimension: the physical cost of sensor data collection itself. Sensors consume nontrivial energy, bandwidth, and processing resources even when their outputs are ultimately down-weighted by the perception model.

\begin{figure}[t]
    \centering
    \includegraphics[width=\linewidth]{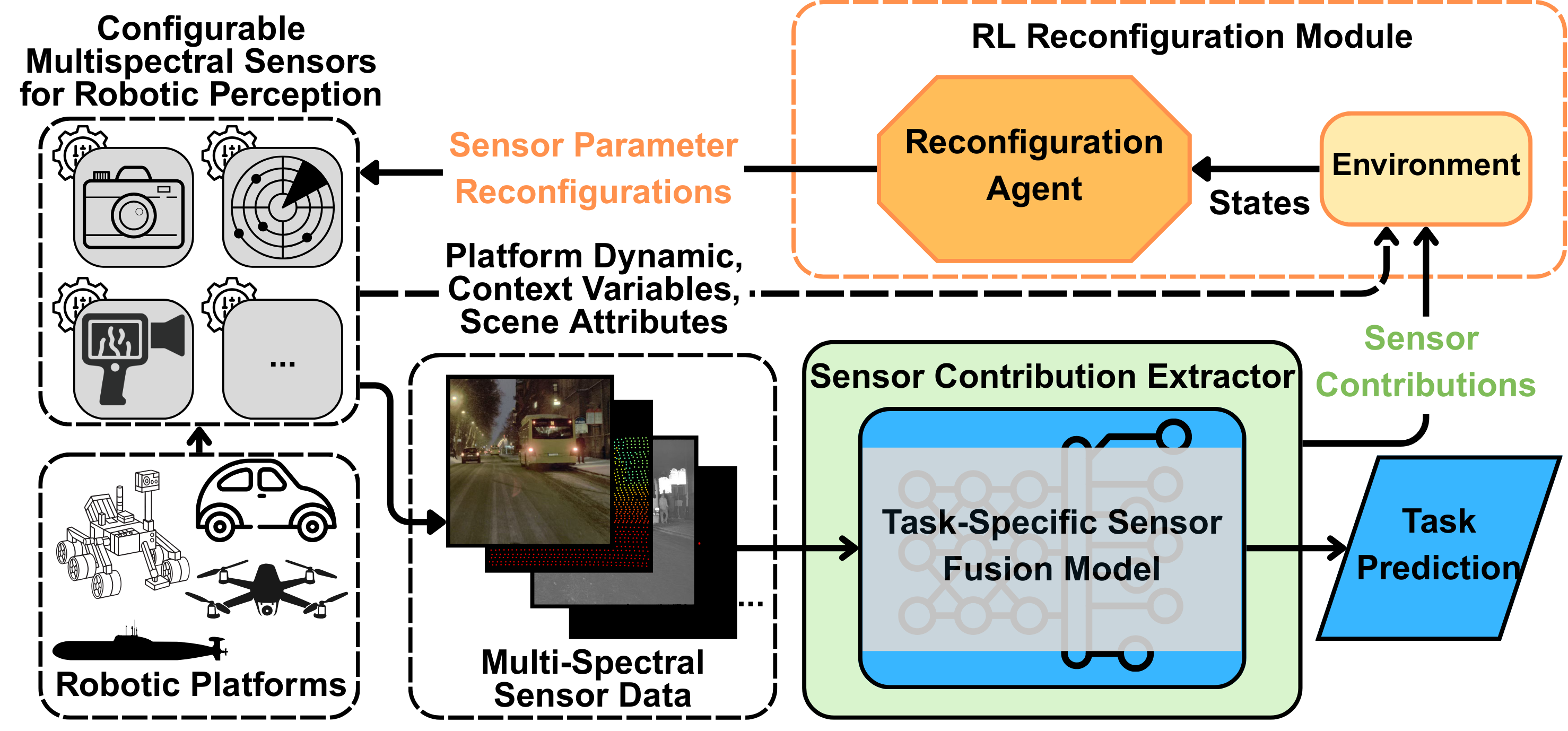}
    \caption{Overview of the proposed \projectname{} framework. A configurable suite of heterogeneous sensors (e.g., RGB, IR, LiDAR, radar) feeds data into a task-specific fusion model wrapped with a contribution extractor, which produces both task predictions and per-modality contribution scores. These scores, combined with platform dynamics and scene attributes, define the state for an RL-based reconfiguration agent. The agent then adjusts sensor parameters in real time (e.g., sampling rate, resolution, sensing range), enabling adaptive perception that maintains accuracy while improving efficiency across different tasks and robotic platforms.}
    \label{fig:teaser}
\end{figure}

In this work, we present an task-adaptable framework that closes this gap by linking contribution-aware fusion directly to real-time control of sensor operation. Specifically, we design and prototype a task-specific detection model that extracts middle layer features from multispectral inputs, estimates each modality’s contribution to the output, and passes these contribution scores as part of the state to an RL agent. The RL agent dynamically tunes sensor configurations such as sampling frequency, resolution, and sensing range. Less informative sensors can be operated at reduced fidelity, while more critical sensors are allocated higher-quality settings. Importantly, once a previously low-contribution sensor begins to provide valuable evidence (e.g., radar returns during fog), the RL agent increases its configuration parameters to reflect its renewed importance.

This work’s novelty lies in unifying sensing, learning, and actuation into a complete reconfiguration loop. Unlike prior approaches that remain at the level of computational fusion or re-weighting, our framework closes the loop by coupling perception with physical sensor control (Figure~\ref{fig:teaser}). Specifically, we are, to our knowledge, the first to bring together:
(\textit{1}) the extraction of quantitative contribution scores for each modality from a unified detection backbone;
(\textit{2}) the integration of these contributions into an RL-based decision-making agent; and
(\textit{3}) the direct adjustment of real-world sensor sampling policies in real time.
It is the combination of these components that enables adaptive sensing with both high detection accuracy and efficiency.

We validate our framework on a mobile rover platform equipped with RGB, IR, mmWave, and depth sensors, demonstrating that adaptive control of sensor configurations significantly reduces sensing and computational load while maintaining, and in some cases improving, perception performance across diverse environmental conditions.

Our contributions are as follows:
\begin{itemize}
    \item We propose \textbf{\projectname{}}, a  framework that closes the loop between multispectral perception and real-time sensor control in dynamic environments.
    \item We present a prototype implementation that instantiates this framework with a detection backbone and reinforcement learning--based reconfiguration agent, deploying this prototype on a mobile rover platform (\emph{\textbf{SPEC}}) to pilot the framework in realistic conditions.
    \item Through experiments, we show that adaptive reconfiguration can reduce sensing and computational cost while preserving, and sometimes improving, perception accuracy, demonstrating the viability of the approach.
\end{itemize}

%% file: sections/2-related.tex
\section{Related Work}

\subsection{Active and Dynamic Multi-Sensor Perception}
The concept of active perception, where robots adjust their sensing strategies to optimize task performance, has deep roots in robotics research~\cite{atanasov_information_2013, atanasov_decentralized_2015, zhang_fisher_2020, placed_survey_2023}. Early works emphasized sensor selection and scheduling to handle occlusion and resource limitations~\cite{xu_adaptive_2024}. For instance, Cho et al.\cite{cho_cocoon_2024} developed an urban driving perception system that dynamically leveraged camera, LiDAR, and radar inputs to improve moving-object detection. Qu et al.\cite{qu_active_2017} proposed a multimodal system with visible, infrared, and hyperspectral sensors that coordinated real-time data collection for target tracking. More recently, robotics platforms have incorporated adaptive sensing pipelines to handle adverse weather or cluttered environments~\cite{palladin_samfusion_2025, park_resilient_2025}. However, these systems typically rely on heuristic rules or static switching policies rather than learning-based approaches that continuously adapt in real time.

\subsection{Computational Fusion and Sensor Importance}
A large body of work has focused on fusion architectures that infer the relative importance of different modalities during perception~\cite{broedermann_condition-aware_2024, zhou_improving_2020, liu_real_2023, zhang_sparselif_2024, zhang_tfdet_2025}. CNN-based methods learn modality importance via gating, confidence weighting, or uncertainty estimation, often yielding robust fusion without transformers. In RGB–thermal detection, CNNs learn per-pixel or per-feature gates that up- or down-weight each stream (e.g., illumination-aware gates, sigmoid point-wise conv gates, or selective-kernel blocks), improving all-day pedestrian detection on open-source datasets such as KAIST and LLVIP~\cite{zhou_improving_2020, xin_multi-modal_2024, hwang_multispectral_2015, jia_llvip_2023}. For LiDAR–camera, CNN pipelines (Siamese/backbone fusion and depth-image projections) estimate reliability and merge cues directly in feature space, boosting detection under complementary failure modes~\cite{liu_real_2023}. Recent LiDAR–camera detectors further introduce uncertainty-aware fusion modules inside CNN stacks to quantify per-modality trust and condition the fusion weights accordingly~\cite{zhang_sparselif_2024}. In RGB-T, modern CNN designs remain competitive by targeting pedestrian-specific cues with target-aware fusion and video extensions~\cite{zhang_tfdet_2025}.

Closely related are gating and selection approaches that suppress or deactivate unreliable sensors. Gated Modality Selection Networks~\cite{cao_multi-modal_2023} have been proposed for real-time monitoring tasks, dynamically reducing the influence of noisy or missing modalities. Other approaches formalize modality selection under resource constraints, selecting an optimal subset of sensors for each inference step~\cite{he_ecient_2024}. While these strategies improve efficiency and interpretability, they remain confined to the feature-processing stage, re-weighting or discarding sensor data after collection but do not reduce the cost of data acquisition itself.

\subsection{Robust Fusion Under Missing Sensors}
Another line of work seeks to ensure perception remains robust when some modalities are absent~\cite{wu_deep_2024}. Approaches such as UniBEV~\cite{wang_unibev_2024} and Immortal~\cite{park_resilient_2025} show that object detectors can be trained to perform well across varying subsets of available sensors. These models use strategies like modality dropout or Mixture-of-Modal-Experts routing to gracefully handle missing data~\cite{park_resilient_2025}. Such robustness is a critical prerequisite for systems that dynamically toggle sensor usage. Our work complements these efforts by providing the control mechanism that decides when and how sensors should be used, enabling real systems to exploit robustness in practice.

\subsection{Reinforcement Learning for Sensor Configuration}
Reinforcement learning (RL) has been widely explored for sensor scheduling and active perception, formulating sensor configuration as a sequential decision-making problem~\cite{kwok_reinforcement_2004, bartolomei_semantic-aware_2021, cheng_reinforcement_2018, xu_applr_2020}.  In both wireless sensor networks and robotic platforms, RL agents have been used to decide which sensors to activate in order to balance task performance with energy consumption~\cite{heo_predictive_2022, shurrab_iot_2022}. To improve robustness, \cite{liu_learning_2017} proposed “sensor dropout” during training in deep RL, while \cite{braun_attention-driven_2019} developed attention-driven policies that selectively focus on informative sensors in an end-to-end manner. These approaches highlight the potential of RL for sensor management but typically frame the problem as a binary on/off activation. More recent efforts have extended RL toward non-binary control~\cite{heo_predictive_2022, huang_context-aware_2025}, yet they remain limited. Existing methods rarely adjust continuous sensing parameters (e.g., sampling frequency or resolution) during ongoing tasks, nor do they exploit importance scores derived from task-specific fusion backbones to guide RL decisions.

%% file: sections/3-method.tex
\begin{figure*}[!t]
    \centering
    \includegraphics[width=0.8\textwidth]{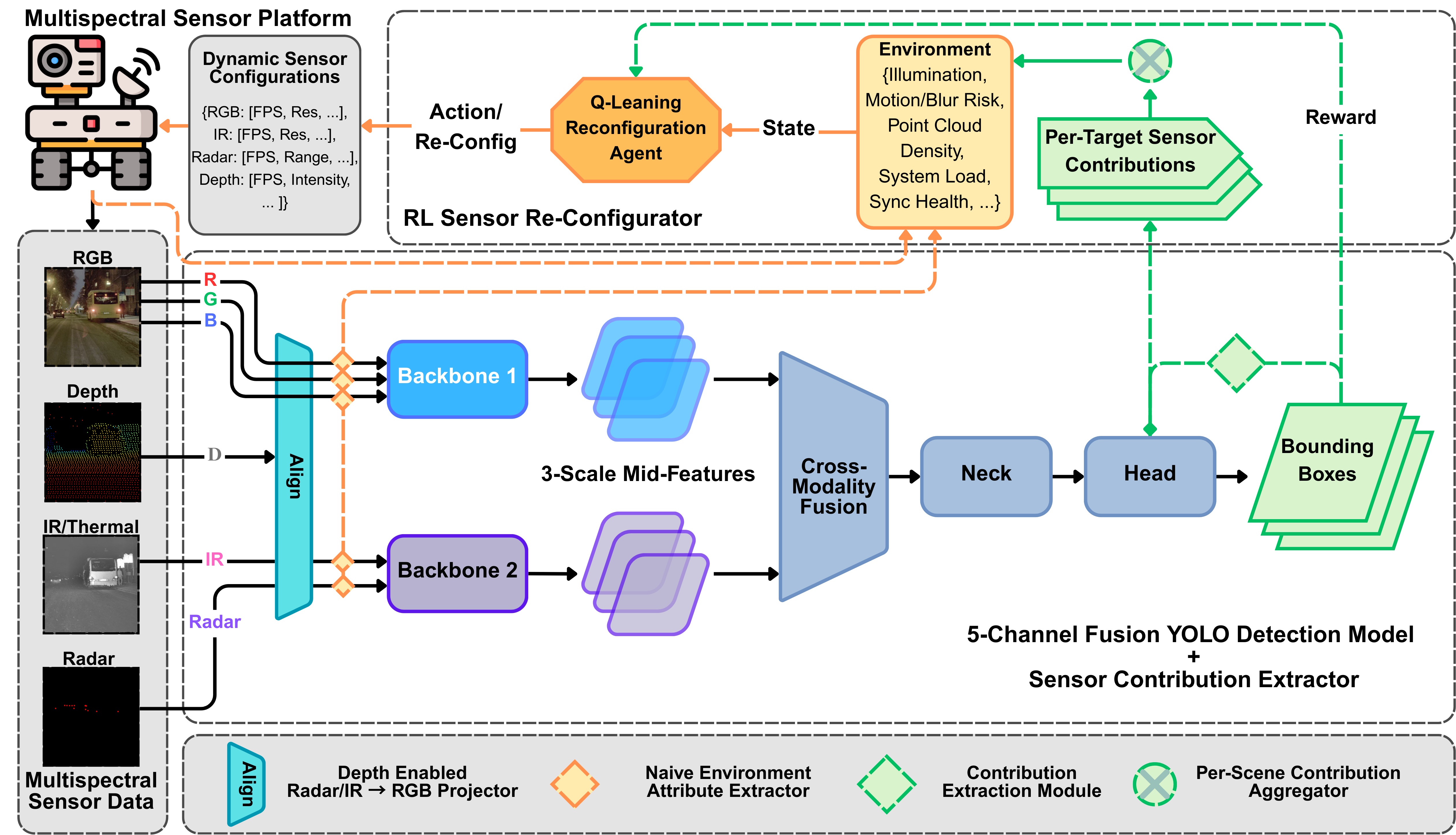}
    \caption{Implementation of the \projectname{} framework for detection tasks. A configurable suite of heterogeneous sensors (RGB, IR, radar, depth) provides synchronized inputs to a detection backbone composed of two lightweight feature extractors and a mid-level fusion module. A sensor contribution extractor produces per-target and scene-level contribution scores, which are combined with environment and system attributes to define the state of an RL-based reconfiguration agent. The agent dynamically adjusts sensor parameters in real time, closing the loop between perception and sensing control.}
    \label{fig:system_architecture}
\end{figure*}

\section{Method}
In this section, we first present an overview of our task-adaptable online sensor reconfiguration framework (\projectname{)}, highlighting its key components and closed-loop design. We then describe the implementation details of our prototype on a multispectral rover platform (\emph{SPEC}), where the framework is instantiated with a detection-oriented backbone and evaluated under diverse operating conditions.

\subsection{\projectname{} Framework}
We propose \projectname{}, a task-adaptable multispectral sensor reconfiguration framework that tightly couples perception with real-time sensing control (Figure~\ref{fig:teaser}). The framework begins with a configurable suite of heterogeneous sensors (e.g., RGB, infrared, LiDAR, radar) that provide raw multispectral data streams. These data are fed into a task-specific fusion backbone wrapped with a sensor contribution extractor, which not only produces task predictions (e.g., object detection) but also generates quantitative contribution scores that reflect the relative utility of each modality for the current task and environment.

A key feature of \projectname{} is that these contribution scores are not derived from fixed heuristics or manually designed weighting rules. Instead, they are emergent signals learned implicitly by the fusion backbone, trained over diverse conditions and tasks. This allows the framework to capture subtle, context-dependent patterns of sensor importance that would be difficult to encode explicitly, and ensures that reconfiguration policies benefit from broad, data-driven generalization.

To enable closed-loop adaptation, these learned contribution scores are combined with platform dynamics (e.g., velocity, motion state), scene attributes (e.g., lighting, density, clutter), and task context variables to define the state for a reinforcement learning (RL)-based reconfiguration agent. The RL agent evaluates this state and outputs optimized sensor configurations in real time, dynamically adjusting parameters such as sampling frequency, resolution, and sensing range. In practice, this means that sensors deemed less informative are down-sampled or temporarily deactivated, while those identified as critical under current conditions are allocated higher fidelity.

By linking fusion-derived importance with direct control of sensor hardware, \projectname{} creates a closed reconfiguration loop that adapts sensing policies on the fly. This design allows robotic platforms to maintain or even improve perception accuracy under diverse operating conditions, while simultaneously reducing redundant data collection and computational overhead. Unlike prior works that stop at feature-level re-weighting, \projectname{} explicitly bridges perception and actuation, enabling energy- and bandwidth-efficient perception across tasks and robotic platforms.

\subsection{Modality Contribution Extraction for Detection}
We now describe our detection-oriented fusion backbone and the mechanism used to extract modality contribution scores (see bottom of Figure~\ref{fig:system_architecture}).

Recent multi-sensor fusion architectures generally follow two main directions: transformer-based cross-attention mechanisms and CNN-based early, mid, or late fusion. Since our goal is efficient, on-device inference, we adopt a CNN-based design in which each modality is projected to a 2D representation before being processed by two lightweight detection backbones. This approach enables explicit quantification of modality contributions while maintaining real-time performance. Our detection model builds on a 5-channel extension of the YOLO family~\cite{redmon_you_2016}, which natively supports bounding box prediction and confidence estimation.

The first stage of the framework extracts mid-level features from multispectral inputs, including RGB, infrared (IR), mmWave radar, and depth. RGB inputs are processed independently by a dedicated convolutional backbone (Backbone 1). IR, mmWave radar, and depth inputs are first projected and spatially aligned with the RGB view through a 3D-to-2D transformation module. These aligned 2D maps are then processed jointly by a shared convolutional backbone (Backbone 2).

The outputs of these branches form modality-specific mid-level features, which are then unified through a lightweight fusion block. The fused representation is passed through the neck and head layers of the YOLO-based architecture to produce task-specific outputs (e.g., bounding boxes). To estimate modality importance, a contribution extraction module quantifies per-target contributions by back-propagating gradient signals to each modality channel. These per-target values are aggregated into scene-level contribution scores, which form part of the state input to the RL agent.

\textbf{Backbone design.} Both backbones follow the YOLOv8~\cite{jocher_ultralytics_2023} architecture, incorporating CSP-based convolutional stages with spatial pyramid pooling for efficient multi-scale representation learning. We employ mid-level fusion, concatenating modality-specific features prior to the detection head.

\textbf{Cross-modal alignment.} To enable pixel-level correspondence, all modalities are explicitly calibrated to the RGB reference frame. Extrinsic calibration parameters are estimated between each modality and the RGB camera, and depth maps are used as a geometric anchor. Both 2D modalities (IR/thermal, gated imaging) and 3D modalities (LiDAR, mmWave radar) are reprojected into the RGB plane, ensuring spatial alignment for fusion.

\textbf{Contribution metric.} Contribution scores are computed via a gradient-based attribution function. For a given target bounding box, gradients are back-propagated, and channel-wise contributions are aggregated over the bounding box region. This produces explicit, object-level importance estimates for each sensing modality, which can then be summarized into scene-level scores for adaptive reconfiguration.

\subsection{RL for Real-time Sensor Reconfiguration}
\label{sec:sec:rl}
To dynamically adapt sensor configurations, we introduce an RL-based reconfiguration module. The environment state comprises both scene attributes and system-level signals, including platform velocity, object density, and modality contribution scores. The agent employs a tabular Q-learning framework~\cite{watkins_q-learning_1992}, chosen for its efficiency and tractability in real-time settings, to optimize sensor parameters such as frame rate, resolution, and sensing range.  

At each timestep, the system aggregates modality contributions and environmental variables to compute a reward signal. This reward incentivizes the agent to prioritize sensors that provide higher utility for the current task while reducing reliance on redundant or costly sensing. Based on this signal, the RL agent outputs reconfiguration actions that adjust the multi-sensor suite online.  

\textbf{Synchronization and Action Space.} All active modalities are synchronized to a global cap of 30 Hz and min of 1 Hz. We set the lower bound to 1 fps, rather than 0, to ensure periodic sampling so that a modality can be reintroduced when its contribution becomes useful. In principle, the effective rate could be lower (e.g., one frame every 3 s), but we preserve occasional sampling to maintain adaptability. Resolution settings further influence algorithmic latency, computational load, and downstream detection quality. The action space is defined as a set of discrete resolution per modality:  
\begin{itemize}
    \item RGB: $\{1280{\times}720,\, 960{\times}540,\, 640{\times}360\}$  
    \item Thermal: $\{160{\times}120,\, 320{\times}240\}$  
    \item mmWave: \{range-prioritized resolution, velocity-prioritized resolution\}
\end{itemize}
The mmWave modality differs from RGB and thermal in that the agent selects between two preference modes rather than fixed resolution parameters. Each preference mode corresponds to a different radar configuration trade-off: when ``prefer range resolution'' is selected, the system prioritizes fine-grained distance measurements; when ``prefer velocity resolution'' is selected, it prioritizes accurate motion detection. Given the selected preference and the current FPS constraint, the system uses a predefined lookup table to determine the four actual mmWave parameters: Range Resolution (m), Maximum Unambiguous Range (m), Maximum Radial Velocity (m/s), and Radial Velocity Resolution (m/s). This lookup approach is necessary because mmWave radar parameters are interdependent—the chirp duration determined by FPS creates inherent trade-offs between range and velocity measurement capabilities that cannot be independently specified like pixel resolutions in visual modalities.

\textbf{State Representation.}  
We design a compact tabular state representation by discretizing six observable factors, each computed over a short sliding window ($w \in [0.5, 1.0]$\,s). This temporal smoothing mitigates momentary fluctuations. Formally, the state at time $t$ is defined as
\[
s_t \in \mathcal{S}
= \mathcal{I} \times \mathcal{B} \times \mathcal{D} \times \mathcal{U} \times \mathcal{Y} \times \mathcal{C},
\]
where each factor is discretized into a small set of ordinal bins (e.g., \{\emph{low}, \emph{medium}, \emph{high}\}). This yields a tractable state space of size $|\mathcal{S}| \leq 3^6 = 729$, suitable for tabular Q-learning.  

The six factors are:  
\begin{itemize}
    \item \textbf{Illumination ($\mathcal{I}$).} Estimated from normalized image intensity statistics or histogram coverage; discretized into \{\emph{low}, \emph{medium}, \emph{high}\}.  
    \item \textbf{Motion ($\mathcal{B}$).} Combines commanded/sensed velocities with image-space proxies; discretized into \{\emph{low}, \emph{medium}, \emph{high}\}.  
    \item \textbf{mmWave point-cloud density ($\mathcal{D}$).} Derived from radar point count, SNR, or CFAR trigger rate; discretized into \{\emph{sparse}, \emph{medium}, \emph{dense}\}.  
    \item \textbf{System load ($\mathcal{U}$).} Measured from GPU/CPU utilization or per-frame latency; discretized into \{\emph{low}, \emph{medium}, \emph{high}\}.  
    \item \textbf{Synchronization health ($\mathcal{Y}$).} Captures timestamp jitter and stale-frame ratio relative to $f_{\mathrm{sync}}$; discretized into \{\emph{good}, \emph{fair}, \emph{poor}\}.  
    \item \textbf{Detection confidence ($\mathcal{C}$).} Aggregates detection confidences using  
    \[
    \mathrm{ConfAgg}_t = \frac{1}{K_t}\sum_{k=1}^{K_t} c_{(k),t}, 
    \qquad K_t = \min\{3,\,N_t\},
    \]  
    where $c_{(k),t}$ is the $k$-th largest detection confidence and $N_t$ is the number of detections. If $N_t=0$, we set $\mathrm{ConfAgg}_t=0$. The aggregated score is discretized into \{\emph{low}, \emph{medium}, \emph{high}\} using thresholds $(\tau_\ell,\tau_h)$.  
\end{itemize}

\textbf{Reward Design.}  
At each step, the reward balances perception gains against system costs:
\begin{equation}
\label{eq:reward}
r_t = \alpha \cdot \Delta \text{Quality}_t
- \beta \cdot P_t
- \gamma \cdot L_t
- \delta \cdot \mathbb{1}[a_t \neq a_{t-1}],
\end{equation}
where $P_t$ is the instantaneous power consumption and $L_t$ the system latency.  

The quality term is defined as the clipped change in detection confidence:
\begin{equation}
\label{eq:dq}
\Delta \text{Quality}_t = \operatorname{clip}\!\big(\mathrm{ConfAgg}_t - \mathrm{ConfAgg}_{t-1},\, -\kappa,\, \kappa\big).
\end{equation}
Here, $\alpha, \beta, \gamma, \delta \ge 0$ are tunable weights, and the last term penalizes frequent configuration switching to encourage stability.

\subsection{SPEC Rover Platform}

\begin{figure}[t]
    \centering
    \includegraphics[width=0.9\columnwidth]{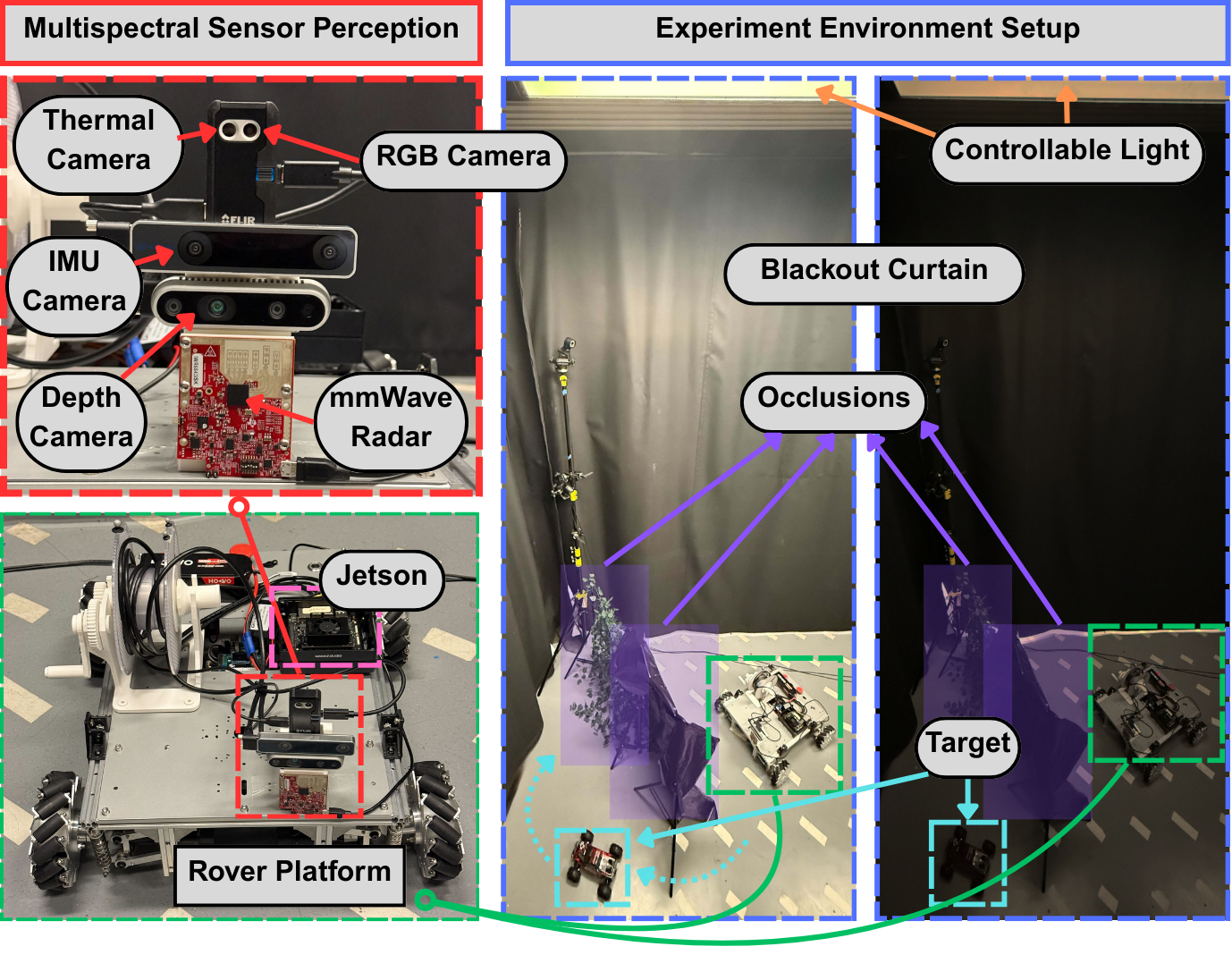}
    \caption{\emph{SPEC} rover platform equipped with a multispectral sensor perception module (RGB, thermal, depth, IMU, and mmWave radar) and the experimental environment. The environment includes controllable lighting and occlusion elements to emulate challenging real-world deployment scenarios, with target trajectories defined for evaluation.}
    \label{fig:platform}
\end{figure}

To validate the framework, we deploy it on a rover-based platform, \emph{SPEC}, equipped with multispectral sensors, including RGB, IR, mmWave radar, and depth cameras. \emph{SPEC} streams synchronized inputs to the task-specific model while the RL reconfigurator runs online to adjust sensing policies.

Figure~\ref{fig:platform} shows the experimental platform for physical deployment.
The platform consists of a 541$\times$581mm mobile rover equipped with four mecanum wheels for omnidirectional mobility. The system is powered by an NVIDIA Jetson Orin Nano edge computer, providing 67 TOPS of AI performance with 8GB of unified memory for real-time inference and RL policy updates.

The multi-modal sensing suite includes: (1) a FLIR One Pro thermal imaging system providing synchronized RGB and thermal IR imagery, (2) a Texas Instruments IWR6843ISK mmWave radar operating at 60-64 GHz with configurable chirp parameters for range-Doppler-angle measurements, (3) an Intel RealSense D435i depth camera with a depth range of 0.3-3m, and (4) an Intel RealSense T265 tracking camera with dual fisheye sensors and IMU for platform dynamics monitoring.

%% file: sections/4-experiment.tex
\section{Experiments and Analysis}
We evaluate the proposed framework through a combination of controlled experiments and dataset-driven analysis, aiming to assess both perception performance and system efficiency under diverse conditions. First, we validate that the task-model–derived contribution scores meaningfully capture the relative utility of each sensing modality under different environmental conditions, using open datasets. Second, we conduct controlled experiments on the \emph{SPEC} rover platform to assess end-to-end system performance. These in-lab scenarios examine whether the adaptive reconfiguration agent can preserve or improve perception quality while simultaneously reducing sensing and computational costs compared to static and heuristic baselines.

\subsection{Contribution Estimation Validation}
\begin{figure}[t]
    \centering
    \includegraphics[width=\columnwidth]{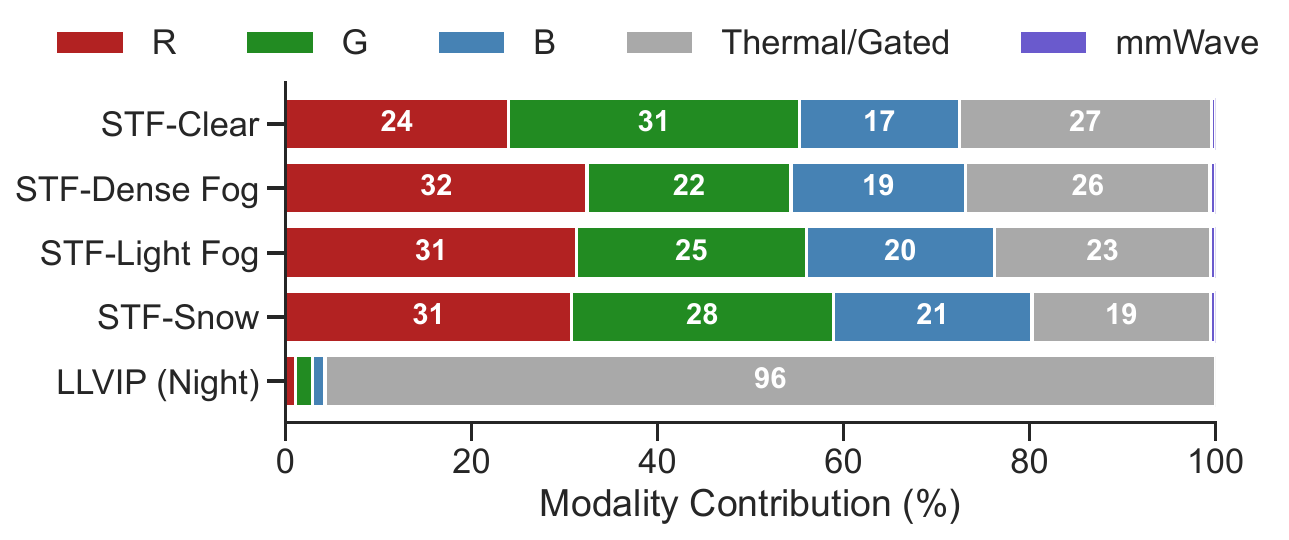}
    \caption{Per-modality contribution scores across diverse open datasets under varying environmental conditions.}
    \label{fig:contribution_map}
\end{figure}

\begin{figure*}[t]
    \centering
    \includegraphics[width=\linewidth]{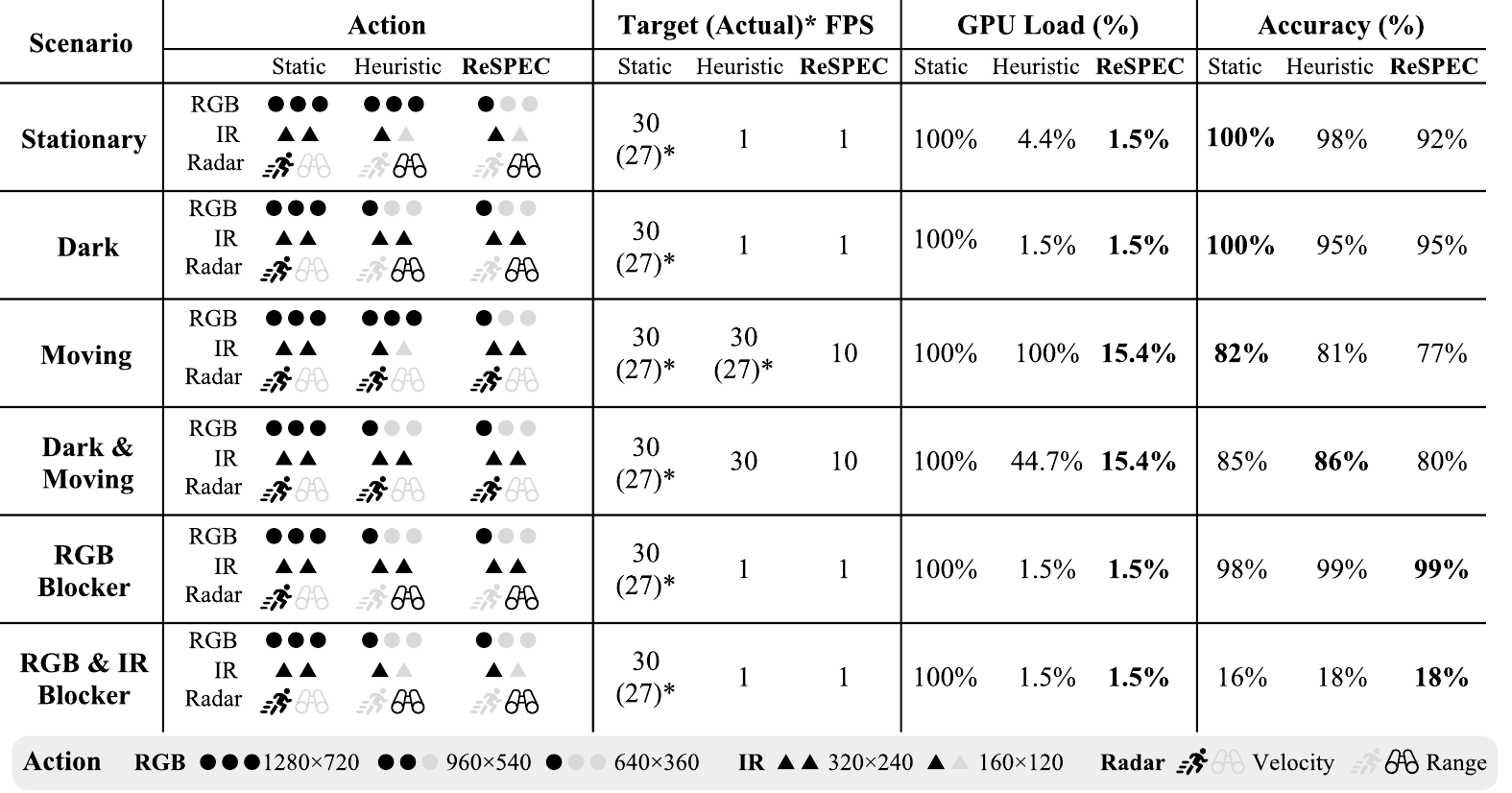}
    \caption{End-to-end evaluation of \projectname{} on the \emph{SPEC} rover platform across controlled in-lab scenarios with varying lighting, target motion, and occlusion. The figure compares adaptive reconfiguration with static (all sensors on) and heuristic (rule-based) baselines, reporting the stabilized actions selected by our system after iterative updates. Results are shown in terms of chosen actions within the defined action space, detection accuracy, and system load. (The FPS cap was 27 instead of 30, as the Jetson Orin Nano reached full load at 27 FPS.)}
    \label{fig:e2e_eval}
\end{figure*}


To verify that task-model–derived contribution scores capture sensor utility, we analyze per-modality contributions across diverse environmental conditions using open datasets SeeingThroughFog (STF)~\cite{bijelic_seeing_2020} and LLVIP~\cite{jia_llvip_2023}. As summarized in Figure~\ref{fig:contribution_map}, the three RGB channels remain the primary drivers on the STF sequences (Clear, Dense Fog, Light Fog, Snow), accounting for roughly 70–80\% of the contribution, while the Thermal/Gated modality provides a consistent secondary share ($\approx 19–27\%$); mmWave contributes negligibly in these scenes. In contrast, LLVIP (Night) shows IR dominance ($\approx 96\%$) with minimal RGB contribution. This is consistent with detection outcomes on the LLVIP validation split: 8,268 detections with IR+RGB versus 8,380 when RGB is intentionally suppressed. The learned scores mirror this trend, assigning $[0.011, 0.019, 0.013, 0.958]$ to (R, G, B, IR), respectively. Together, these results indicate that learned contribution scores track environment- and dataset-dependent utility, justifying their use as a guiding signal for reconfiguration.

\subsection{End-to-End Testing on the \emph{SPEC} Rover}




We designed a series of controlled scenarios in our in-lab experimental environment (Figure~\ref{fig:platform}) to systematically evaluate the robustness of our system. By intentionally altering lighting conditions, moving the target object, and introducing different forms of occlusion (Figure~\ref{fig:e2e_eval}), we tested the end-to-end performance and efficiency of our adaptive framework against static and heuristic sensor configurations.

Specifically, we considered the following scenarios: (1) \textit{Good lighting, static environment}: Serves as a baseline to evaluate performance under ideal conditions. (2) \textit{Poor lighting, static environment}: Tests whether the system can emphasize sensors more resilient to illumination changes (e.g., IR or radar). (3) \textit{Good lighting, moving target}: Evaluates adaptation to dynamic scenes where temporal resolution and sampling rate become critical. (4) \textit{Poor lighting, moving target}: Represents a challenging condition requiring simultaneous adaptation across modalities (e.g., boosting IR and radar while reducing RGB reliance). (5) and (6) \textit{Good lighting with occlusions}: Assesses whether the system can reallocate sensing resources to modalities less affected by partial visibility (e.g., IR or mmWave radar).

These scenarios were executed sequentially to mimic the evolving conditions a robot may encounter in real-world deployments. Our goal was to demonstrate that the adaptive reconfiguration agent can rapidly adjust sensing strategies as environments shift.

To ensure fair comparison, we report the actions taken by our system after the policy has converged—that is, once iterative updates stabilize and the selected configuration no longer changes. These stabilized actions are then compared against static (all sensors on) and heuristic (rule-based) baselines in terms of their positions within the defined action space, as well as the resulting detection accuracy and system load.

On the \emph{SPEC} rover platform, adaptive reconfiguration yielded measurable benefits: it reduced average GPU computational load by 29.3\% with only 5.3\% accuracy degradation compared to the heuristic baseline. In addition, accuracy was preserved—and in some cases improved—because the agent prioritized the most informative modalities under each condition.

Figure~\ref{fig:e2e_eval} presents the actions produce by our system, showing that our RL-based policy can discover optimal operating points. For example, in low-light conditions with a moving target, the RL agent increased IR sampling while down-sampling RGB, achieving accuracy close to the full static configuration but with significantly lower compute and bandwidth cost.

Taken together, these results underscore the practicality of scaling adaptive sensing to embedded robotic platforms with tight resource budgets. By dynamically adjusting to changing environmental conditions, our system achieves a balanced tradeoff between efficiency and perception quality, paving the way for broader deployment of resource-aware multi-sensor robotics.



%% file: sections/5-conclusion.tex
\section{Discussion and Limitations}
Our results demonstrate the feasibility of closing the loop between perception and sensor control, enabling robotic platforms to dynamically adapt sensing policies while maintaining robust task performance. Beyond the immediate detection use case, this framework has broader implications for multi-sensor robotic perception and control.

\textbf{Broader Applicability Across Tasks.} Although instantiated here with a detection-oriented backbone, the framework is inherently task-agnostic. The same closed-loop principle, deriving modality contributions and feeding them into an RL reconfigurator, can be extended to other perception and decision-making tasks. For instance, in multi-object tracking, contribution scores could be used to dynamically prioritize sensors that best maintain target identity under occlusion. In planning and navigation, reconfiguration policies could down-sample sensors in open, uncluttered spaces and increase fidelity in dense, dynamic environments. Similarly, in SLAM and mapping, adaptive sensing could conserve resources by lowering LiDAR resolution in static scenes while increasing it during loop closures or in geometrically ambiguous areas. These extensions highlight the potential of the framework as a general foundation for adaptive sensing across diverse robotic pipelines.

\textbf{Multi-Agent and Networked Scenarios.} Our experiments focused on a single robotic platform, but adaptive sensing is especially relevant in multi-robot systems where communication bandwidth is a shared, limited resource. Extending the framework to coordinate sensing policies across multiple agents would enable collaborative perception while minimizing redundant sensing and transmission. Similarly, in network-constrained environments, the reconfiguration policy could integrate communication costs directly into the reward function, trading off local fidelity with shared situational awareness.

\textbf{Limitations.} This work has several limitations that should be acknowledged. The framework relies on the stability of task-model-derived contribution estimates, yet noisy or unstable attributions could mislead the reconfiguration policy. In practice, hardware-level switching delays also introduce latency in sensor activation and deactivation, which constrains how quickly the system can respond to sudden environmental changes. Moreover, our current design does not explicitly encode strict safety or latency guarantees, leaving the possibility of unsafe sensing gaps in critical scenarios. Finally, while we employ tabular Q-learning for tractability in our prototype, scaling to more complex tasks may require more advanced reinforcement learning methods, such as deep RL, with careful attention to sample efficiency and stability. Future work should address these challenges by incorporating uncertainty-aware and safety-constrained RL, investigating wake-on-event or compress-then-sense paradigms for ultra-low-power sensing, extending evaluation to broader tasks such as planning and SLAM, and exploring coordinated reconfiguration strategies in multi-robot and communication-limited settings.

\section{Conclusion}
We introduced \projectname{}, a closed-loop framework that unifies sensing, learning, and actuation for multispectral robotic perception. Unlike prior methods that restrict adaptivity to the feature space, our approach directly reconfigures sensor operation in real time, tuning sampling frequency, resolution, and sensing range based on learned modality contributions and reinforcement learning policies. Experiments on the \emph{SPEC} rover platform demonstrated that this design substantially reduces sensing and computational overhead while maintaining, and in some cases improving, perception performance across diverse conditions. These results highlight the practical benefits of linking “what the model attends to” with “what the robot senses next.” More broadly, this work points toward a future of adaptive, resource-aware perception in robotics. By embedding intelligence at the sensing layer, robots can dynamically align their data acquisition strategies with task demands and environmental conditions. This principle provides a general blueprint for building efficient, resilient, and scalable multi-sensor systems, applicable not only to detection but also to tracking, navigation, and multi-robot collaboration.